\PassOptionsToPackage{unicode}{hyperref}
\PassOptionsToPackage{hyphens}{url}
\documentclass[
  10pt,
  letterpaper,
]{article}
\usepackage{lmodern}
\usepackage{amssymb,amsmath}
\usepackage{ifxetex,ifluatex}
\ifnum 0\ifxetex 1\fi\ifluatex 1\fi=0 
  \usepackage[T1]{fontenc}
  \usepackage[utf8]{inputenc}
  \usepackage{textcomp} 
\else 
  \usepackage{unicode-math}
  \defaultfontfeatures{Scale=MatchLowercase}
  \defaultfontfeatures[\rmfamily]{Ligatures=TeX,Scale=1}
\fi
\IfFileExists{upquote.sty}{\usepackage{upquote}}{}
\IfFileExists{microtype.sty}{
  \usepackage[]{microtype}
  \UseMicrotypeSet[protrusion]{basicmath} 
}{}
\makeatletter
\@ifundefined{KOMAClassName}{
  \IfFileExists{parskip.sty}{%
    \usepackage{parskip}
  }{
    \setlength{\parindent}{0pt}
    \setlength{\parskip}{6pt plus 2pt minus 1pt}}
}{
  \KOMAoptions{parskip=half}}
\makeatother
\usepackage{xcolor}
\IfFileExists{xurl.sty}{\usepackage{xurl}}{} 
\IfFileExists{bookmark.sty}{\usepackage{bookmark}}{\usepackage{hyperref}}
\hypersetup{
  pdftitle={Pre-Flight: A Benchmark for Evaluating Large Language Models on Aviation Operational Knowledge},
  pdfauthor={Alex Brooker (Airside Labs, London, United Kingdom); Tim Hughes (Mahino Research, Christchurch, New Zealand)},
  hidelinks,
  pdfcreator={LaTeX via pandoc}}
\usepackage[margin=1in]{geometry}
\usepackage{longtable,booktabs}
\usepackage{etoolbox}
\makeatletter
\patchcmd\longtable{\par}{\if@noskipsec\mbox{}\fi\par}{}{}
\makeatother
\IfFileExists{footnotehyper.sty}{\usepackage{footnotehyper}}{\usepackage{footnote}}
\makesavenoteenv{longtable}
\usepackage{graphicx}
\makeatletter
\def\maxwidth{\ifdim\Gin@nat@width>\linewidth\linewidth\else\Gin@nat@width\fi}
\def\maxheight{\ifdim\Gin@nat@height>\textheight\textheight\else\Gin@nat@height\fi}
\makeatother
\setkeys{Gin}{width=\maxwidth,height=\maxheight,keepaspectratio}
\makeatletter
\def\fps@figure{htbp}
\makeatother
\usepackage{textcomp}
\ifdefined\DeclareUnicodeCharacter \DeclareUnicodeCharacter{2020}{\dag}
\DeclareUnicodeCharacter{00B7}{\textperiodcentered}
\DeclareUnicodeCharacter{2248}{\ensuremath{\approx}} \fi

\title{Pre-Flight: A Benchmark for Evaluating Large Language Models on
Aviation Operational Knowledge}
\author{Alex Brooker (Airside Labs, London, United Kingdom) \and Tim
Hughes (Mahino Research, Christchurch, New Zealand)}
\date{29 June 2026 \textperiodcentered{} Licensed under CC BY 4.0}

\begin{document}
\maketitle

\hypertarget{abstract}{%
\subsection{Abstract}\label{abstract}}

Large language models (LLMs) are increasingly proposed for aviation
business operations, from documentation and training generation to
customer facing assistants. General purpose benchmarks do not measure
whether a model reasons safely and correctly about aviation specific
operational knowledge, and the high stakes, regulated nature of the
domain makes that gap consequential. We present Pre-Flight, an open
source benchmark of 300 multiple choice questions drawn from
international standards and airport ground operations material, covering
international airport ground operations, ICAO and US FAA regulations,
aviation general knowledge and complex operational scenarios. Questions
were authored and reviewed by practitioners with experience in air
traffic management, ground operations and commercial flying. We evaluate
a range of contemporary commercial and open weight models using the
Inspect evaluation framework, scoring by accuracy under a standard
multiple choice protocol, and we maintain the leaderboard on a rolling
basis as new models are released. Against an informal expert reference
of around 95\%, obtained from a low sample quiz of aviation
professionals at a conference, even the strongest model evaluated
(released in 2026) reaches 82.7\%, having improved only gradually from
roughly 75\% in early 2025. A substantial and persistent gap below
expert level reliability therefore remains. We release the dataset, the
evaluation harness and the results, and the benchmark is available
within the community evaluations package distributed with
\texttt{inspect\_evals}. We argue that domain specific evaluation of
this kind is a necessary precondition for responsible deployment of
generative AI in non safety critical aviation operations.

\emph{Headline figures are taken from the leaderboard snapshot of 29
June 2026 (Table 2).}

\begin{center}\rule{0.5\linewidth}{0.5pt}\end{center}

\hypertarget{introduction}{%
\subsection{1. Introduction}\label{introduction}}

Aviation business operations span several subdomains, each with distinct
and demanding characteristics. Network planning processes large volumes
of demographic and market data to optimise routes and schedules. Airport
operations require continuous, forward looking resource balancing in
real time. Asset management ranges from short term maintenance tracking
to multidecade fleet lifecycle analysis. Airspace and procedure design
integrates many disparate data formats with operational expertise under
complex regulatory constraints.

What makes these subdomains particularly challenging is their
interconnected nature and the heterogeneity of their data. Information
encoded in legacy formats such as NOTAMs and Type B messages over AFTN
is used alongside modern digital services. Maintenance records persist
as scanned paper with wet signatures and stamps. Considerable effort
over the past two decades has gone into rationalising aviation
information models, notably System Wide Information Management (SWIM)
under ICAO's Global Air Navigation Plan, but industry uptake has been
slow and migration will take decades. The result is a landscape in which
simple message structures from the 1960s coexist with heavyweight
geospatial XML schemas, and in which a large share of operationally
relevant knowledge, including regulations, training material and
procedures, remains unstructured text.

This combination of structured and unstructured information is precisely
where generative AI appears promising. The motivating question is
whether LLMs can bring these sources together to answer operational
questions that were previously intractable, and, critically, whether we
can trust the answers. Many aviation decisions are high stakes, whether
because of the financial sums involved or the need for accuracy, and
errors can carry reputational and operational consequences. General
purpose evaluation does not tell us whether a model genuinely
understands aviation operations or merely produces plausible sounding
output.

This paper makes three contributions. First, we motivate the need for
aviation specific LLM evaluation and situate it against general and
other domain specific benchmarks. Second, we introduce Pre-Flight, an
open source multiple choice benchmark of aviation operational knowledge,
and describe its construction and composition. Third, we report results
for a set of contemporary models and provide a qualitative analysis of
characteristic failure modes. We restrict our scope to non safety
critical business operations; generative AI is not proposed here for
safety critical functions, consistent with the assurance led,
incremental approach to AI in aviation set out in the FAA's Roadmap for
Artificial Intelligence Safety Assurance {[}FAA, 2024{]}.

\begin{center}\rule{0.5\linewidth}{0.5pt}\end{center}

\hypertarget{background-and-related-work}{%
\subsection{2. Background and related
work}\label{background-and-related-work}}

\hypertarget{what-a-good-llm-evaluation-requires}{%
\subsubsection{2.1 What a good LLM evaluation
requires}\label{what-a-good-llm-evaluation-requires}}

A useful evaluation framework should be authentic, in that it reflects
real world performance using representative data; unbiased; attentive to
safety and robustness, including resilience to adversarial input; and
where possible informative about the explainability of outputs. Standard
automatic metrics each capture only part of this picture: perplexity for
language modelling, BLEU {[}Papineni et al., 2002{]} and METEOR
{[}Banerjee and Lavie, 2005{]} for translation quality, and ROUGE
{[}Lin, 2004{]} for summarisation overlap. For multiple choice knowledge
tasks of the kind used here, accuracy under a fixed protocol is the
standard and most interpretable measure.

A recurring difficulty is that strong aggregate scores can mask shallow
competence. Models frequently express high confidence that is not
matched by accuracy, a calibration gap documented in recent factuality
work {[}Wei et al., 2024{]}. In a domain whose safety culture depends on
a precise understanding of system limitations, this overconfidence is a
specific liability, and it implies that factual recall measured in
isolation is insufficient.

\hypertarget{public-benchmarks-domain-benchmarks-and-contamination}{%
\subsubsection{2.2 Public benchmarks, domain benchmarks and
contamination}\label{public-benchmarks-domain-benchmarks-and-contamination}}

General multitask benchmarks such as MMLU {[}Hendrycks et al., 2021{]}
measure broad knowledge but include little aviation operational content
and do not test the integration of regulation, procedure and operational
context. Domain specific benchmarks have become established in other
regulated fields, including medicine {[}MedQA; Jin et al., 2021{]} and
law {[}LegalBench; Guha et al., 2023{]}, precisely because general
benchmarks fail to capture the knowledge and failure modes that matter
for deployment in those settings. Aviation has lacked an equivalent open
resource.

Public benchmarks also face contamination: because models are trained on
large web corpora that may include the benchmark itself, public scores
can be inflated and may overstate generalisation {[}Sainz et al.,
2023{]}. This motivates holding out a private test partition, an
approach used in other domains to preserve discriminative power as
models improve.

\hypertarget{domain-specific-evaluation-in-practice}{%
\subsubsection{2.3 Domain specific evaluation in
practice}\label{domain-specific-evaluation-in-practice}}

Regulated industries already deploy LLMs and vision models under domain
specific evaluation: clinical summarisation and decision support in
healthcare, fraud detection and risk assessment in finance, defect
detection in manufacturing. The common thread is that predeployment
evaluation, and continued in service monitoring, are tailored to the
domain's data structures, error costs and regulatory requirements.
Pre-Flight applies the same logic to aviation operational knowledge.

\hypertarget{aviation-and-aerospace-llm-benchmarks}{%
\subsubsection{2.4 Aviation and aerospace LLM
benchmarks}\label{aviation-and-aerospace-llm-benchmarks}}

Two recent aerospace efforts are closest to our work, and both are
complementary to it, sitting at different layers of the aviation stack.
The Aerospace Language Understanding Evaluation (ALUE), from the FAA and
MITRE, targets the systemic layer, the National Airspace System and air
traffic management, and provides a configurable framework for assessing
whether models genuinely understand aviation language and context, with
a roadmap toward multimodal and retrieval grounded tasks such as chart
extraction and consulting operational manuals {[}Mangortey et al.,
2025{]}. PilotBench evaluates LLMs as agents on safety critical flight
trajectory and attitude prediction from real general aviation telemetry,
scored on a composite of regression accuracy and instruction and safety
adherence {[}Wu et al., 2026{]}. Pre-Flight occupies a distinct,
complementary niche at the operational layer: it measures declarative
operational and regulatory knowledge (ground operations, ICAO procedures
and dispatch) through a single, fixed, openly released multiple choice
dataset with a public rolling leaderboard distributed in
\texttt{inspect\_evals}, and it deliberately restricts its scope to non
safety critical business operations. Where ALUE provides flexible
assurance infrastructure for the airspace system and PilotBench probes
physics governed prediction, Pre-Flight supplies a concrete, reusable
measure of the under the wing operational knowledge that commercial
deployments depend on. Together they signal aviation's move toward
domain specific LLM evaluation from complementary directions.

\begin{center}\rule{0.5\linewidth}{0.5pt}\end{center}

\hypertarget{the-pre-flight-benchmark}{%
\subsection{3. The Pre-Flight
benchmark}\label{the-pre-flight-benchmark}}

\hypertarget{task-and-format}{%
\subsubsection{3.1 Task and format}\label{task-and-format}}

Pre-Flight is a multiple choice benchmark. Each item presents a question
with four or five answer choices and exactly one correct answer. The
benchmark targets understanding of ICAO annex documentation, flight
dispatch rules, and airport ground operations safety procedures and
protocols, with questions drawn from international airline and airport
ground operations safety manuals. Items are grouped by source section
into the five categories used throughout this paper, namely
international airport ground operations, ICAO rules and regulations, FAA
rules and regulations, aviation trivia, and complex ground scenarios,
whose full composition is given in Table 1 (Section 3.3).

\hypertarget{dataset-construction}{%
\subsubsection{3.2 Dataset construction}\label{dataset-construction}}

The benchmark items are drawn from authoritative aviation source
material: international airport Ground Operations Safety Manuals (GOSM),
United States regulations (14 CFR), ICAO Annexes, and general aviation
knowledge documentation. Items were authored by the lead author together
with a small group of aviation practitioners with experience in air
traffic management, ground operations and commercial flying. Each item
is stored in JSONL format with a unique identifier, the question stem,
four or five answer options, and a single target answer.

The current public release is an ``easy'' tier. A separate, harder tier
is under development and is deliberately withheld to preserve
discriminative power as models improve; it is not yet released.

Correct answers are grounded in the cited source documents, where the
gold answer for each item can be located directly. Answers were
partially and manually validated by the lead author and Tim Hughes, both
with around 25 years of aviation experience, checking items against the
source documents and their domain knowledge, and a further aviation
subject matter expert reviewed a subset of the items. Validation was
therefore expert based and source grounded, but partial: not every item
received independent second review, and no formal interannotator
agreement statistic was computed.

\hypertarget{dataset-composition}{%
\subsubsection{3.3 Dataset composition}\label{dataset-composition}}

The benchmark comprises 300 multiple choice items across five categories
(Table 1). The set is deliberately weighted towards operational ground
content: international airport ground operations is by far the largest
category, and together with the ICAO and FAA regulatory categories it
accounts for 96\% of the benchmark. Two categories, complex ground
scenarios and aviation trivia, are small and are best read as
indicative.

\textbf{Table 1. Composition of the Pre-Flight benchmark.}

\begin{longtable}[]{@{}lrr@{}}
\toprule
Category & Items & Share\tabularnewline
\midrule
\endhead
International airport ground operations & 152 & 50.7\%\tabularnewline
ICAO rules and regulations & 85 & 28.3\%\tabularnewline
FAA rules and regulations & 51 & 17.0\%\tabularnewline
Aviation trivia & 8 & 2.7\%\tabularnewline
Complex ground scenarios & 4 & 1.3\%\tabularnewline
\textbf{Total} & \textbf{300} & \textbf{100\%}\tabularnewline
\bottomrule
\end{longtable}

This same five way grouping is used for the category analysis in Section
5.2.

\hypertarget{availability-and-licensing}{%
\subsubsection{3.4 Availability and
licensing}\label{availability-and-licensing}}

The public dataset is released under the MIT licence as
\texttt{AirsideLabs/pre-flight-06} on Hugging Face, and the benchmark is
included in the community evaluations package distributed with
\texttt{inspect\_evals} (\texttt{UKGovernmentBEIS/inspect\_evals}) {[}UK
AI Security Institute et al., 2024b{]}, where it can be run via
\texttt{inspect\ eval\ inspect\_evals/pre\_flight}. Inclusion in
\texttt{inspect\_evals} means the evaluation was accepted as a community
contribution to a repository maintained by the UK AI Security Institute;
it does not imply endorsement of the benchmark by that body. The
benchmark has been part of the \texttt{inspect\_evals} collection since
March 2025, and the dataset has accrued 11,416 downloads on Hugging Face
as of June 2026, indicating active community uptake. A separate harder
tier is in development and is not publicly released, providing a
contamination resistant complement to the public set.

\begin{center}\rule{0.5\linewidth}{0.5pt}\end{center}

\hypertarget{experimental-setup}{%
\subsection{4. Experimental setup}\label{experimental-setup}}

We evaluate using the Inspect framework {[}UK AI Security Institute,
2024a{]}, running the benchmark as the \texttt{pre\_flight} task
distributed in \texttt{inspect\_evals} {[}UK AI Security Institute et
al., 2024b{]}, which loads the public \texttt{AirsideLabs/pre-flight-06}
test split at its pinned revision (\texttt{439d2d1}). The task pairs
Inspect's standard \texttt{multiple\_choice} solver with its
\texttt{choice} scorer, and the reported metric is accuracy, the
proportion of items answered correctly. Items are scored by
deterministic exact match of the model's selected option against the
gold answer; no model graded or LLM judge scoring is used. The results
record two scorer labels, accuracy and choice, but for single answer
multiple choice both reduce to the same exact match comparison, so
results do not depend on the scoring components that differ across
Inspect framework versions.

We ran the task in its default configuration. Prompting is zero shot,
using the solver's default multiple choice template with no custom
system prompt, and no generation parameters (temperature, top-p or
maximum tokens) were overridden, so each model used its provider's or
runtime's default sampling settings. The task sets no epoch count, so
each model is evaluated once over the full 300 item set (epochs = 1);
the error reported in Table 2 is therefore the binomial standard error
for n = 300 (approximately 0.025). The 300 item dataset is text only;
the visual probes described in Section 6 are separate, illustrative
material and are not part of the scored benchmark.

Rather than a one off batch, models are evaluated on a rolling basis as
they are released, and the leaderboard is maintained continuously. The
results reported here are a snapshot taken on 29 June 2026, spanning
models released from early 2024 to mid 2026. Thirteen earlier models are
shown at their original scores and have not yet been rerun on the
current dataset revision (\texttt{439d2d1}); they are marked with a
dagger in Table 2. Commercial models were accessed through their
providers' APIs (OpenAI, Anthropic, Google) and through Groq; selected
open weight models, including quantised variants, were run locally on
DGX Spark hardware. Where the same model was served under more than one
configuration it is listed separately and labelled accordingly in Table
2, distinguishing provider APIs, Groq, and locally served quantised
variants. The leaderboard records each model's public release date
rather than the date on which it was evaluated; accordingly, release
dates, not evaluation dates, are used in Figure 1 and Table 2.

\begin{center}\rule{0.5\linewidth}{0.5pt}\end{center}

\hypertarget{results}{%
\subsection{5. Results}\label{results}}

\hypertarget{overall-results}{%
\subsubsection{5.1 Overall results}\label{overall-results}}

Table 2 reports accuracy for all 44 evaluated models against an informal
expert reference of around 95\%. Figure 1 plots accuracy against model
release date and traces the frontier over time. Three findings stand
out. First, the gap to expert level performance is large and persistent:
the strongest model in the snapshot, GPT-5.5, reaches 82.7\%, around
twelve points below the informal expert reference. Second, progress has
been slow relative to that gap: the frontier rose quickly through 2024,
from the mid 60s to the low 70s, then continued to climb but more
gradually, adding roughly eight points across 2025 and into 2026 to
reach 82.7\%, still rising, but slowly, and well short of expert
reliability. Third, open weight models are now competitive at the
frontier: a locally run, quantised Qwen3.5 122B variant reaches 77.3\%,
behind only the four leading API systems (GPT-5.5, GPT-5, Claude Opus
4.8 and Gemini 2.5 Pro) and ahead of GPT-5.1, with several open models
clustered among the leaders. We also note a clear outlier: ALLaM 2 7B
scores 25.3\%, at or near chance for the four to five choice format,
illustrating that a model strong in other settings can collapse entirely
on this domain.

\includegraphics[width=0.92\textwidth,height=\textheight]{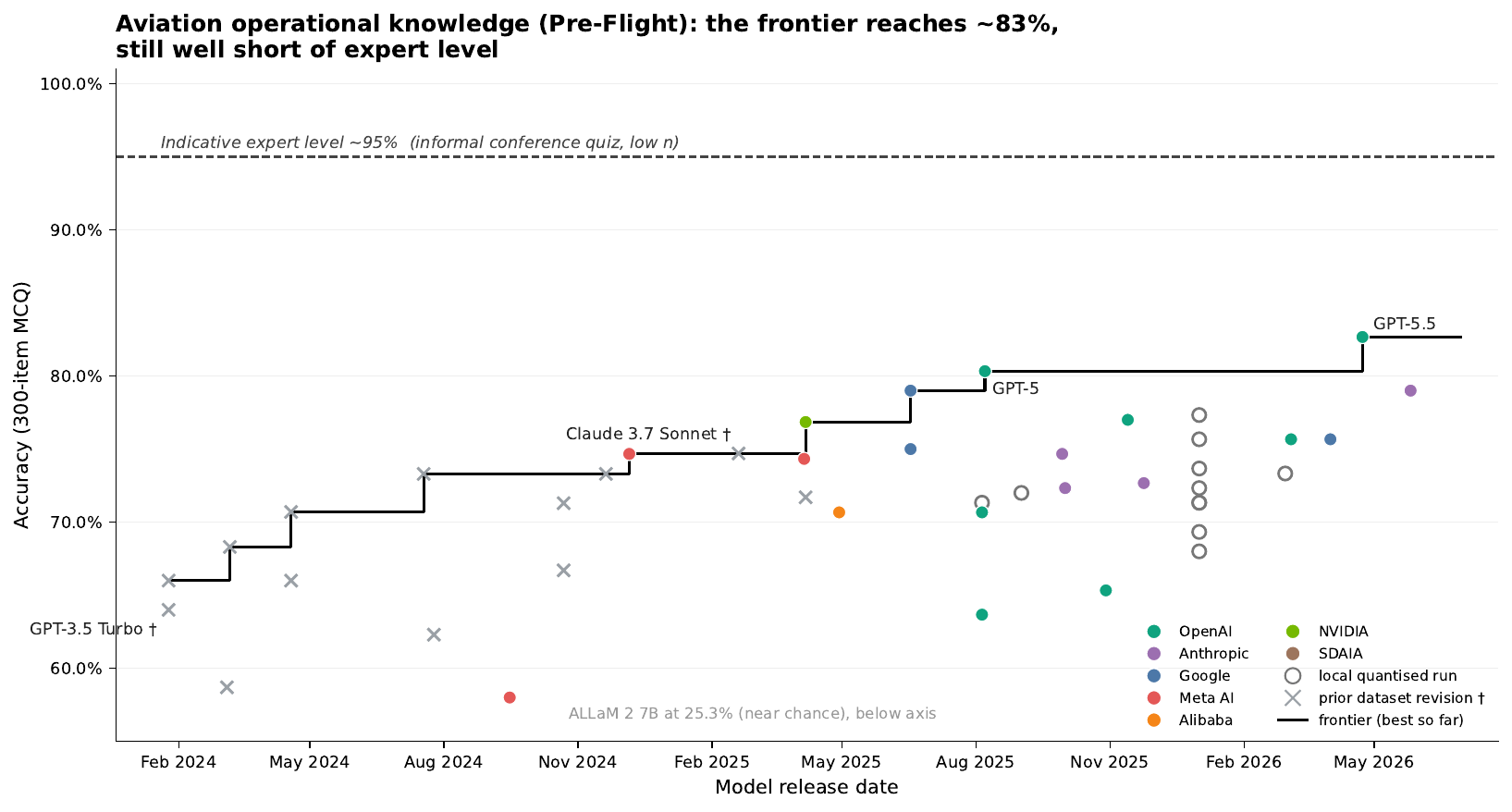}

\textbf{Figure 1.} Pre-Flight accuracy by model release date; the
horizontal axis is each model's public release date, not its evaluation
date. Filled circles are API served models, coloured by organisation;
hollow circles are open weight models run locally on DGX Spark hardware
(approximate dates). Grey crosses (†) are earlier models shown at their
original scores, not yet rerun on the current dataset revision, so the
frontier line mixes revisions before mid 2025 and is indicative there.
The dashed line is the informal expert reference. ALLaM 2 7B (25.3\%)
falls below the plotted range.

\textbf{Table 2. Pre-Flight accuracy by model (snapshot, dataset
revision \texttt{439d2d1}, 29 June 2026; n = 300; 44 models; informal
expert reference \textasciitilde95\%).} Error is the binomial standard
error for n = 300; models marked ``local Spark'' were run locally on DGX
Spark hardware. The 13 models marked † are shown at their original
scores and have not yet been rerun on the current revision.

\begin{longtable}[]{@{}llrr@{}}
\toprule
Model & Organisation & Accuracy & SE\tabularnewline
\midrule
\endhead
GPT-5.5 & OpenAI & 82.7\% & 0.022\tabularnewline
GPT-5 & OpenAI & 80.3\% & 0.023\tabularnewline
Claude Opus 4.8 & Anthropic & 79.0\% & 0.024\tabularnewline
Gemini 2.5 Pro & Google & 79.0\% & 0.024\tabularnewline
Qwen3.5 122B-A10B int4 (local 2x Spark) & Alibaba & 77.3\% &
0.024\tabularnewline
GPT-5.1 & OpenAI & 77.0\% & 0.024\tabularnewline
NVIDIA Llama 3.1 Nemotron & NVIDIA & 76.8\% & 0.024\tabularnewline
GPT-5.4 & OpenAI & 75.7\% & 0.025\tabularnewline
Gemini 3.5 Flash & Google & 75.7\% & 0.025\tabularnewline
Qwen3.6 35B-A3B FP8 (local Spark) & Alibaba & 75.7\% &
0.025\tabularnewline
Gemini 2.5 Flash & Google & 75.0\% & 0.025\tabularnewline
Claude 3.7 Sonnet † & Anthropic & 74.7\% & 0.025\tabularnewline
Claude 4.5 Sonnet & Anthropic & 74.7\% & 0.025\tabularnewline
Llama 3.3 70B & Meta AI & 74.7\% & 0.025\tabularnewline
Llama 4 Scout 17B & Meta AI & 74.3\% & 0.025\tabularnewline
Qwen3-Coder-Next FP8 (local Spark) & Alibaba & 73.7\% &
0.025\tabularnewline
Gemma 4 26B-A4B (local Spark) & Google & 73.3\% & 0.026\tabularnewline
GPT-4o (Nov 2024) † & OpenAI & 73.3\% & 0.026\tabularnewline
GPT-4o Mini † & OpenAI & 73.3\% & 0.026\tabularnewline
Claude Sonnet 4.6 & Anthropic & 72.7\% & 0.026\tabularnewline
Claude Haiku 4.5 & Anthropic & 72.3\% & 0.026\tabularnewline
MiniMax M2.7 AWQ (local 2x Spark) & MiniMax & 72.3\% &
0.026\tabularnewline
Qwen3.5 35B-A3B FP8 (local Spark) & Alibaba & 72.0\% &
0.026\tabularnewline
Llama 4 Scout † & Meta AI & 71.7\% & 0.026\tabularnewline
GPT-OSS 120B MXFP4 (local Spark) & OpenAI & 71.3\% &
0.026\tabularnewline
Nemotron-3 Super 120B-A12B NVFP4 (local Spark) & NVIDIA & 71.3\% &
0.026\tabularnewline
Qwen3-Coder-Next int4 (local Spark) & Alibaba & 71.3\% &
0.026\tabularnewline
Claude 3.5 Sonnet (Oct 2024) † & Anthropic & 71.3\% &
0.026\tabularnewline
Llama 3 70B (Groq) † & Meta AI & 70.7\% & 0.026\tabularnewline
GPT-OSS 120B & OpenAI & 70.7\% & 0.026\tabularnewline
Qwen3 32B & Alibaba & 70.7\% & 0.026\tabularnewline
DiffusionGemma 26B-A4B NVFP4 (local Spark) & Google & 69.3\% &
0.027\tabularnewline
Claude 3 Haiku † & Anthropic & 68.3\% & 0.027\tabularnewline
Nemotron-3 Nano 30B-A3B (local Spark) & NVIDIA & 68.0\% &
0.027\tabularnewline
Claude 3.5 Haiku † & Anthropic & 66.7\% & 0.027\tabularnewline
GPT-4 Preview (Jan 2024) † & OpenAI & 66.0\% & 0.027\tabularnewline
Llama 3 8B (Groq) † & Meta AI & 66.0\% & 0.027\tabularnewline
GPT-OSS Safeguard 20B & OpenAI & 65.3\% & 0.027\tabularnewline
GPT-3.5 Turbo † & OpenAI & 64.0\% & 0.028\tabularnewline
GPT-OSS 20B & OpenAI & 63.7\% & 0.028\tabularnewline
Gemma 2 9B † & Google & 62.3\% & 0.028\tabularnewline
Qwen QWQ 32B † & Alibaba & 58.7\% & 0.028\tabularnewline
Llama 3.1 8B & Meta AI & 58.0\% & 0.028\tabularnewline
ALLaM 2 7B & SDAIA & 25.3\% & 0.025\tabularnewline
\textbf{Informal expert reference} & n/a &
\textbf{\textasciitilde95.0\%} & n/a\tabularnewline
\bottomrule
\end{longtable}

\hypertarget{item-level-and-category-analysis}{%
\subsubsection{5.2 Item level and category
analysis}\label{item-level-and-category-analysis}}

The following analysis is computed over the 31 model cohort scored on
dataset revision \texttt{439d2d1}, and characterises that cohort rather
than every entry in the rolling leaderboard.

Difficulty is unevenly distributed. Of 300 items, 9 (3.0\%) were
answered incorrectly by every model, 24 (8.0\%) by 80 to 99\% of models,
39 (13.0\%) by 50 to 79\%, and the remaining 228 (76.0\%) by fewer than
half. By category, US (FAA) regulations are by far the hardest, at
53.0\% mean failure (n = 51), followed at a distance by complex ground
scenarios (33.9\%, n = 4), aviation trivia (30.2\%, n = 8), ICAO rules
and regulations (23.5\%, n = 85) and international airport ground
operations (22.7\%, n = 152); the micro averaged failure rate across all
items is 28.4\%. The two smallest categories rest on very few items (n =
4 and n = 8) and are high variance. The broad pattern is clear: models
handle widely published international safety and ICAO content well and
struggle most with US regulatory detail, and eight of the nine items
failed by every model are US regulation items. A plausible contributor
is that the sector's open documentation culture has placed much of the
international material in model training data, which would also help
explain why even small, older models score above 70\% overall.

Separating recall from reasoning, 60.3\% of items require reasoning
rather than direct recall, and models fared somewhat worse on these
(30.1\% failure versus 25.9\% for recall, a gap of 4.1 points). The most
common reasoning demands are comparison, operational reasoning and
scenario analysis, with a smaller set requiring explicit calculation.
This is consistent with the broader claim that aggregate knowledge
scores overstate operational competence: difficulty concentrates where
multistep reasoning is required, which is also where a deployed system
would most benefit from retrieval augmentation and tool use rather than
parametric knowledge alone.

One caveat shapes how the US regulation result should be read.
Inspecting the most failed items showed that some were answer key errors
rather than model failures: on these the models converge almost
unanimously on an option that matches the governing regulation more
closely than the recorded key. We identified two of these as answer key
errors, the definition of ``operational control'' and the aircraft
dispatcher's duty to brief the pilot in command, and corrected the keys;
with the keys fixed, every model but one answers them and they leave the
hard set. A few of the remaining items show the same signature (for
instance which 14 CFR part 121 document records emergency crewmember
functions, and the runway ``clearway'' versus ``stopway'' distinction)
and are flagged for expert revalidation, so the FAA rate should still be
read as a modest upper bound. This also illustrates a useful method:
strong, consistent disagreement between many independent models and a
key is an efficient way to surface benchmark errors. Where the consensus
answer contradicts the source document, as in the ground operations item
on mandatory safety cone placement, the failure is genuine relative to
that source, though it reflects common practice rather than fabrication.

\begin{center}\rule{0.5\linewidth}{0.5pt}\end{center}

\hypertarget{qualitative-analysis-of-failure-modes}{%
\subsection{6. Qualitative analysis of failure
modes}\label{qualitative-analysis-of-failure-modes}}

Alongside the multiple choice benchmark, we ran a small set of
structured probes to illustrate characteristic failure modes. These are
qualitative and illustrative, not part of the scored dataset, and we
report them as such.

\textbf{Reasoning versus pattern matching.} A simplified variant of the
classic river crossing puzzle exposes overfitting: when presented with a
basic farmer and chicken scenario, several models produced elaborate but
incorrect solutions, pattern matching against the familiar training data
version rather than reasoning from the stated problem. Reasoning
oriented models that review intermediate steps before answering largely
avoided this failure.

\textbf{Physical and temporal constraints.} A snow clearance scheduling
task, in which only one aircraft stand can be cleared at a time and
stands must be cleared between a departure and the next arrival, tested
handling of real world constraints. Models could manage basic scheduling
but sometimes failed to respect all physical and temporal limitations,
selecting orderings that violated stand availability.

\textbf{Complex visual reasoning.} Using a 1970s Aeroflot route map, we
set a three part task: identify the hub and network structure, interpret
a schedule for a specific day of the week, and find an efficient route
between distant cities. Models reliably identified Moscow as the hub but
struggled with day of week schedule interpretation and route
optimisation, which require chaining several inferences across a dense
diagram. The pattern suggests vision language models can extract
structure from complex diagrams but fall short of reliable multistep
interpretation, with implications for any screen watching assistant
role.

The common thread across these probes is that aggregate knowledge scores
do not capture reasoning reliability under real world constraints, which
reinforces the case for evaluation that goes beyond factual recall.

\begin{center}\rule{0.5\linewidth}{0.5pt}\end{center}

\hypertarget{limitations}{%
\subsection{7. Limitations}\label{limitations}}

Several limitations bound the conclusions of this work. The benchmark is
multiple choice, which measures recognition rather than open ended
generation or operational judgement, and high accuracy here should not
be read as fitness for deployment. The questions derive from public
standards and manuals, so training data contamination cannot be excluded
for the public tier; the strong performance of even small, older models
on widely published safety material is consistent with some
contamination. The harder tier under development is intended to mitigate
this. The canonical run is a snapshot in time, and model rankings will
change as new models are released, so results require periodic
re-running to stay current. The expert reference figure is informal: it
derives from a small, self selected sample of conference attendees
answering a question subset under uncontrolled conditions, and should be
read as indicative rather than as a validated human baseline. The
qualitative probes in Section 6 are illustrative and small in number,
and should not be generalised quantitatively. Finally, the benchmark
covers selected operational areas and does not claim full coverage of
aviation business operations, and by design excludes safety critical
functions entirely. The categories are also markedly uneven in size,
from 152 items down to 4, so per category rates for the smallest
categories are noisy and should not be over-interpreted. Answer key
quality is a further limitation: a review of the most failed items found
that several gold answers, concentrated in the US regulation category,
are contestable (on these the models converge on the option that aligns
with the governing regulation), so the FAA failure rate should be read
as an upper bound pending expert revalidation. A small number of items
additionally reference an external figure that is not reproduced in the
text only prompt; rather than treat these as unanswerable, we retain
them deliberately as extremely hard retrieval items, since success
requires the model to infer the question's context and reconstruct a
plausible version of the figure accurately enough to reach the correct
answer.

\begin{center}\rule{0.5\linewidth}{0.5pt}\end{center}

\hypertarget{conclusion-and-future-work}{%
\subsection{8. Conclusion and future
work}\label{conclusion-and-future-work}}

Domain specific evaluation is a precondition for responsible use of
generative AI in aviation business operations. Pre-Flight provides an
open, reproducible measure of aviation operational knowledge, and our
results show that even the strongest current models leave a clear gap
below operational reliability. Future work includes expanding category
coverage, adding open ended and tool augmented tasks, maintaining a
versioned leaderboard against new model releases, and growing the expert
contributor base. We invite the community to run, scrutinise and extend
the benchmark.

\begin{center}\rule{0.5\linewidth}{0.5pt}\end{center}

\hypertarget{acknowledgements}{%
\subsection{Acknowledgements}\label{acknowledgements}}

We thank Andrew Iwanoczko, Professor Achim Czerny and Dr Adrienne
Leonard for reviewing and providing feedback on earlier versions of this
work. We also thank Conor Mullan of Think Research for hosting the
authors and the quiz at the Think Research exhibition stand at Airspace
World 2025, where the aviation professionals whose responses underpin
the informal expert reference answered the benchmark questions.

\begin{center}\rule{0.5\linewidth}{0.5pt}\end{center}

\hypertarget{references}{%
\subsection{References}\label{references}}

Banerjee, S. and Lavie, A. (2005). METEOR: An Automatic Metric for MT
Evaluation with Improved Correlation with Human Judgments. In
\emph{Proceedings of the ACL Workshop on Intrinsic and Extrinsic
Evaluation Measures for Machine Translation and/or Summarization},
pp.~65--72.

Guha, N., Nyarko, J., Ho, D. E., Ré, C., et al.~(2023). LegalBench: A
Collaboratively Built Benchmark for Measuring Legal Reasoning in Large
Language Models. In \emph{Advances in Neural Information Processing
Systems 36 (NeurIPS Datasets and Benchmarks Track)}.

Hendrycks, D., Burns, C., Basart, S., Zou, A., Mazeika, M., Song, D. and
Steinhardt, J. (2021). Measuring Massive Multitask Language
Understanding. In \emph{International Conference on Learning
Representations (ICLR)}.

Jin, D., Pan, E., Oufattole, N., Weng, W.-H., Fang, H. and Szolovits, P.
(2021). What Disease Does This Patient Have? A Large-Scale Open Domain
Question Answering Dataset from Medical Exams. \emph{Applied Sciences},
11(14):6421.

Lin, C.-Y. (2004). ROUGE: A Package for Automatic Evaluation of
Summaries. In \emph{Text Summarization Branches Out}, pp.~74--81.
Association for Computational Linguistics.

Mangortey, E., Singh, S., Chen, S. and Sarkhel, K. (2025). Aviation
Language Understanding Evaluation (ALUE) -- Large Language Model
Benchmark with Aviation Datasets. In \emph{AIAA AVIATION FORUM AND
ASCEND 2025}, paper AIAA 2025-3247 (FAA and MITRE).
https://github.com/mitre/alue

Papineni, K., Roukos, S., Ward, T. and Zhu, W.-J. (2002). BLEU: A Method
for Automatic Evaluation of Machine Translation. In \emph{Proceedings of
the 40th Annual Meeting of the Association for Computational
Linguistics}, pp.~311--318.

Sainz, O., Campos, J. A., García-Ferrero, I., Etxaniz, J., Lopez de
Lacalle, O. and Agirre, E. (2023). NLP Evaluation in Trouble: On the
Need to Measure LLM Data Contamination for each Benchmark. In
\emph{Findings of the Association for Computational Linguistics: EMNLP
2023}, pp.~10776--10787. arXiv:2310.18018.

UK AI Security Institute (2024a). \emph{Inspect AI: Framework for Large
Language Model Evaluations.}
https://github.com/UKGovernmentBEIS/inspect\_ai

UK AI Security Institute, Arcadia Impact and Vector Institute (2024b).
\emph{Inspect Evals: Community-Contributed LLM Evaluations for Inspect
AI.} https://github.com/UKGovernmentBEIS/inspect\_evals

US Federal Aviation Administration (FAA) (2024). \emph{Roadmap for
Artificial Intelligence Safety Assurance.}
https://www.faa.gov/aircraft/air\_cert/step/roadmap\_for\_AI\_safety\_assurance

Wei, J., Karina, N., Chung, H. W., Jiao, Y. J., Papay, S., Glaese, A.,
Schulman, J. and Fedus, W. (2024). Measuring Short-Form Factuality in
Large Language Models. arXiv:2411.04368.

Wu, Y., Liu, H., Li, Z. and Wang, B. (2026). PilotBench: A Benchmark for
General Aviation Agents with Safety Constraints. arXiv:2604.08987.

\begin{center}\rule{0.5\linewidth}{0.5pt}\end{center}

\hypertarget{appendix-a.-consistently-failed-items}{%
\subsection{Appendix A. Consistently failed
items}\label{appendix-a.-consistently-failed-items}}

Nine items were answered incorrectly by every one of the thirty one
models in the cohort: eight US regulation (14 CFR) items and one
international airport ground operations item. We identified two items as
answer key errors, the definition of ``operational control'' and the
party required to brief the pilot in command on weather and facility
irregularities (the aircraft dispatcher), and corrected the keys; with
the keys fixed, every model but one answers them and they leave the set.
A few of the remaining items show the same signature of near unanimous
model disagreement with the key and are candidates for the same review,
for instance which 14 CFR part 121 document records emergency crewmember
functions, and the runway ``clearway'' versus ``stopway'' distinction;
these are flagged for expert revalidation rather than reported as
confirmed model failures. Others are hard because they test a specific
mandate rather than general practice: on the ground operations item on
cone placement, the source manual mandates a cone under the tail on
arrival, with wingtip and other cones placed as recommended by IATA or
as required by airlines, so the near universal model choice (a cone at
each wingtip) describes common ramp practice but not the placement the
manual designates as mandatory. A further four US regulation items just
below the all fail threshold depend on FAA figures (FAA-CT-8080-7D) that
are not supplied in the text only prompt. Rather than treat these as
unanswerable, we retain them as deliberately extreme retrieval items: a
model can succeed only by inferring from the question what the
referenced figure depicts and reconstructing it faithfully enough to
select the correct option, which makes them a demanding test of the
grounded inference that operational deployments require.

\end{document}